\documentclass[a4paper,fleqn]{cas-dc}

\usepackage[numbers]{natbib}
\usepackage{graphicx}
\usepackage{capt-of} 
\def\tsc#1{\csdef{#1}{\textsc{\lowercase{#1}}\xspace}}
\tsc{WGM}
\tsc{QE}
\tsc{EP}
\tsc{PMS}
\tsc{BEC}
\tsc{DE}

\begin{document}
\let\WriteBookmarks\relax
\def\floatpagepagefraction{1}
\def\textpagefraction{.001}

\shorttitle{Large Ontology Models for Enterprise Knowledge Management}


\title [mode = title]{Construct, Align, and Reason: Large Ontology Models for Enterprise Knowledge Management}                      



%
\author[1,2]{Yao Zhang}






\affiliation[1]{organization={Yonyou AI Lab}}

\author[1,2]{Hongyin Zhu}[orcid=0000-0001-5786-7594]

\ead{zhuhongyin@yonyou.com}


\affiliation[2]{organization={Yonyou Network Technology Co., Ltd.},
    country={}
    }



\cortext[cor1]{Corresponding author}



\begin{abstract}
Enterprise-scale knowledge management faces significant challenges in integrating multi-source heterogeneous data and enabling effective semantic reasoning. Traditional knowledge graphs often struggle with implicit relationship discovery and lack sufficient semantic understanding for complex question answering. To address these limitations, we introduce a unified construct--align--reason framework, the large ontology model (LOM). We first build a dual-layer enterprise ontology from structured databases and unstructured text, subsequently fusing these sources into a comprehensive enterprise ontology. To enable instruction-aligned reasoning, we propose a unified three-stage training pipeline: ontology instruction fine-tuning to improve structural understanding; text-ontology grounding to strengthen node semantic encoding; and multi-task instruction tuning on ontology-language pairs with curriculum learning to enhance semantic reasoning and generation. We also construct comprehensive training and evaluation datasets covering diverse ontology reasoning tasks. On this benchmark, our 4B-parameter LOM achieves 89.47\% accuracy and outperforms DeepSeek-V3.2 on complex graph reasoning, indicating effective fusion of ontology structure and language.
\end{abstract}



\begin{keywords}
ontology construction \sep large language model \sep knowledge graph \sep graph encoder 
\end{keywords}

\maketitle

\section{Introduction}
In the wave of digital transformation, enterprise knowledge management faces unprecedented challenges. Traditional systems typically rely on relational databases or basic document management systems to store and manage enterprise knowledge. While these approaches can store structured data effectively, they struggle with complex unstructured knowledge and inter-entity relationships. In recent years, knowledge graphs have emerged as a powerful representation that models enterprise knowledge as graphs, using nodes and edges to capture entities and their relationships, thereby enabling more intelligent querying and reasoning.

Existing approaches for enterprise ontology construction and reasoning face distinct limitations. On the construction side, methods often treat structured databases and unstructured text in isolation. Traditional schema mapping tools \cite{DBLP:conf/cikm/IglesiasJCCV20} struggle to identify implicit relationships (e.g., missing foreign keys) in legacy databases, while standard information extraction models \cite{DBLP:journals/cee/ZhuTZGAND22} lack the domain adaptability to merge these structured backbones with rich textual knowledge. On the reasoning side, a significant semantic gap persists: graph neural networks (GNNs) \cite{DBLP:journals/corr/abs-2403-16033} capture topology but lack the reasoning depth for complex business questions, whereas large language models (LLMs) \cite{DBLP:journals/tkde/PanLWCWW24} possess semantic knowledge but often fail to adhere to rigorous graph structures. This disconnect hampers the ability to perform reliable, multi-hop reasoning over heterogeneous enterprise data. To address these challenges, we present a large ontology model (LOM) that unifies multi-source ontology construction with a structure-aware instruction tuning pipeline, effectively bridging the divide between data integration and semantic reasoning.

Constructing an enterprise ontology is a non-trivial undertaking. We employ a layered approach to build ontologies from both structured databases and unstructured text. For structured databases, where explicit foreign keys are frequently missing, we propose a multi-factor relationship discovery algorithm that analyzes both schema metadata and data content overlap to uncover implicit connections. This enables the construction of a dual-layer ontology comprising an abstract schema layer and a concrete instance layer. For unstructured text, we utilize an LLM\&LOM-based pipeline that performs entity-relation extraction, link prediction, and robust entity disambiguation via a hybrid of symbolic rules and semantic embeddings. Finally, we fuse these heterogeneous sources into a unified enterprise ontology through cross-modal alignment based on tag-description matching.

The second challenge lies in training an LOM capable of deeply understanding and reasoning over these heterogeneous enterprise ontologies. Existing methods often fail to bridge the semantic gap between graph structures and textual knowledge. To address this, we implement a unified three-stage training pipeline. First, we employ ontology instruction fine-tuning to endow the LLM with foundational ontology structural understanding. Second, we introduce a text-ontology grounding stage that trains an alignment projector to fuse textual semantics with ontology features via intra- and inter-type alignment. Finally, we conduct multi-task instruction tuning over ontology-language pairs with curriculum learning, guiding the model from simple predictive tasks to complex generative reasoning. To support this pipeline, we construct a comprehensive CoT-enhanced dataset that captures algorithmic reasoning paths, enabling the model to learn the logic behind ontology-centric operations rather than simple answer mapping.

We conduct systematic evaluation on our datasets, and the 4B-parameter LOM achieves state-of-the-art performance. Our main contributions are:

1. We introduce the large ontology model, a unified construct--align--reason framework that systematically integrates multi-source ontology construction, text-ontology alignment, and instruction-aligned reasoning to resolve the semantic disconnect between structured databases and unstructured knowledge.

2. We design a three-stage instruction alignment training pipeline—comprising ontology instruction fine-tuning, text-ontology grounding, and multi-task instruction tuning over ontology-language pairs—to endow the model with robust complex ontology reasoning capabilities.

3. We construct a comprehensive CoT-enhanced dataset covering 19 diverse graph reasoning tasks. Extensive evaluations demonstrate that our LOM-4B achieves state-of-the-art performance with 89.47\% accuracy on this benchmark, significantly outperforming leading baselines in complex graph reasoning tasks.

\section{Related Work}
\subsection{Ontology Construction from Heterogeneous Data}
Recent studies have revisited ontology construction from relational databases by incorporating agent-based reasoning and large language models to reduce manual schema engineering. Trajanoska et al. \cite{trajanoska2025multi} propose a multi-agent framework in which specialized agents collaboratively perform schema interpretation, mapping rule generation, and semantic alignment between relational tables and knowledge graph schemas. While this approach improves modularity and interpretability, it still relies on explicit coordination protocols and assumes relatively clean schema-level signals, limiting its robustness in large-scale enterprise databases with implicit or noisy relational structures.

More recently, retrieval-augmented and autonomous ontology construction methods have emerged. Nayyeri et al. \cite{DBLP:journals/corr/abs-2506-01232} introduce a RAG-based framework that retrieves schema fragments and instance-level evidence to guide LLM-driven ontology generation, demonstrating improved adaptability across heterogeneous databases. In parallel, AutoSchemaKG \cite{DBLP:journals/corr/abs-2505-23628} explores dynamic schema induction from web-scale corpora, enabling autonomous ontology evolution without predefined schemas. However, these methods primarily focus on either structured databases or unstructured text in isolation, and they do not explicitly address cross-source ontology alignment or the discovery of implicit relational patterns such as hidden foreign keys and enterprise-specific relationship semantics.

A complementary line of research focuses on declarative mapping frameworks for constructing RDF knowledge graphs from structured and heterogeneous data sources. SDM-RDFizer \cite{DBLP:conf/cikm/IglesiasJCCV20} presents an efficient RML interpreter optimized for large-scale RDF generation through parallel execution and standards-compliant mapping, enabling deterministic and reproducible knowledge graph construction. Building upon such infrastructures, Assche et al. \cite{DBLP:conf/kgcw/AsscheJM24} study backward compatibility and rule reuse in RML mapping pipelines, addressing maintainability challenges under evolving schemas. While these systems provide robust engineering solutions for knowledge graph materialization, they rely on manually defined mappings and predefined schemas, and do not address automated ontology induction, implicit relationship discovery, or instruction-aligned reasoning over enterprise knowledge graphs, which are the focus of our work.

\subsection{Ontology Reasoning}
Recent advances in large language models have motivated the integration of graph-structured knowledge into language-centric reasoning frameworks. Early graph language models focus on encoding graph structures into LLM-compatible representations. HiGPT \cite{DBLP:conf/kdd/TangY0SXY024} introduces a heterogeneous graph language model that aligns graph nodes and relations with textual embeddings, enabling joint reasoning over graph topology and semantics. Building upon this paradigm, GraphAgent \cite{DBLP:journals/corr/abs-2412-17029} frames graph reasoning as an agentic process, allowing LLMs to iteratively plan, reason, and interact with graph environments. These approaches demonstrate the potential of agent-based graph reasoning, yet they primarily target generic graph benchmarks and do not explicitly consider the scale, heterogeneity, and dynamic evolution characteristic of enterprise ontologies.

To enhance LLMs’ capability to reason over graph structures, recent work explores instruction tuning and foundation model designs tailored to graph data. GraphInstruct \cite{DBLP:journals/corr/abs-2403-04483} proposes a large-scale instruction dataset covering diverse graph algorithms and demonstrates that instruction-tuned LLMs can acquire non-trivial graph reasoning abilities without explicit graph neural networks. Complementarily, G-Reasoner \cite{DBLP:journals/corr/abs-2509-24276} presents a foundation model for unified reasoning over graph-structured knowledge, emphasizing generalization across tasks and graph modalities. While these methods significantly advance instruction-aligned graph reasoning, they mainly focus on static or synthetic graph settings and provide limited support for grounding graph reasoning in real-world enterprise semantics and heterogeneous knowledge sources.

Another line of research enhances LLM reasoning through explicit interaction with knowledge graphs \cite{DBLP:journals/tkde/PanLWCWW24}. Survey work \cite{DBLP:journals/www/ZhuWCQOYDCZ24} systematically reviews the emerging capabilities of LLMs in KG construction and reasoning, highlighting both opportunities and limitations. On the modeling side, retrieval- and path-based approaches aim to improve multi-hop reasoning. GNN-RAG \cite{DBLP:conf/acl/MavromatisK25} employs graph neural networks to guide retrieval of relevant subgraphs for LLM reasoning, while Zhou et al. \cite{DBLP:conf/acl/ZhouLLLLZWH0025} introduce reflective mechanisms to iteratively refine KG-based reasoning. Paths-over-Graph \cite{DBLP:conf/www/TanWLXYZ25} further explores path-centric reasoning strategies for complex multi-hop queries. In applied settings, Mendes et al. \cite{mendes2024application} demonstrate the feasibility of deploying LLM-powered KGQA systems in enterprise environments. However, these approaches typically treat the knowledge graph as an external tool or retrieval source, rather than jointly modeling ontology construction, graph representation, and instruction-aligned reasoning within a unified framework.

\section{Approach}
In this section, we present the LOM-based construct--align--reason framework, where \emph{construct} denotes ontology construction, \emph{align} denotes ontology--text alignment, and \emph{reason} denotes generative ontology reasoning. We first introduce the model architecture, followed by unified training and inference strategies for complex ontology reasoning. We then detail the construction of our CoT-enhanced graph reasoning dataset. Finally, we describe our ontology construction methodology, which integrates structured databases and unstructured text into a unified enterprise ontology.
\subsection{Large Ontology Model}
\subsubsection{Model Architecture}
\begin{center}
\includegraphics[width=0.45\textwidth]{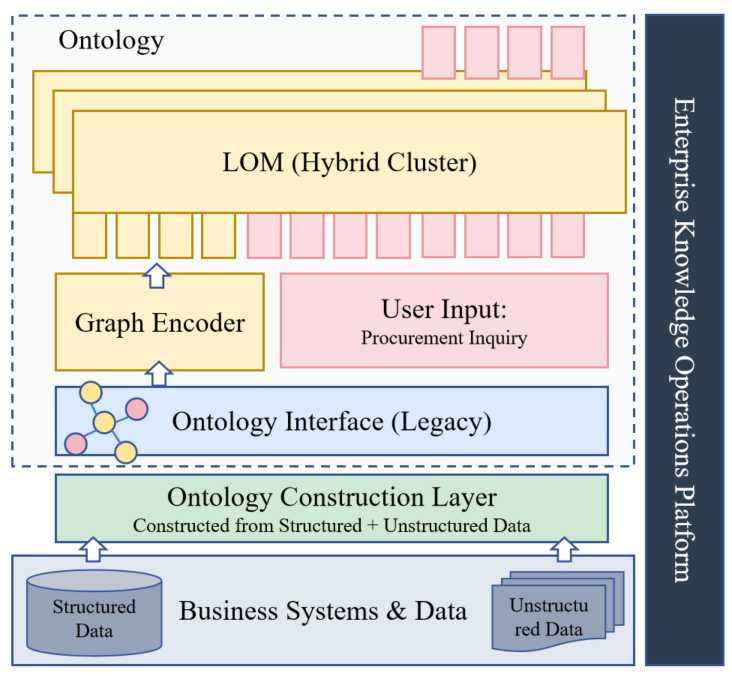}
\captionof{figure}{Model Architecture Overview}
\label{fig.arch}
\end{center}


As illustrated in Figure \ref{fig.arch}, the LOM architecture is situated within the enterprise knowledge operations platform, bridging bottom-layer business systems \& data (structured and unstructured data) with top-layer ontology applications. Specifically, an ontology construction layer is introduced, constructed from structured and unstructured data. Data flows through an ontology interface into the model, which employs a heterogeneous graph representation $\mathcal{G}=(\mathcal{V}, \mathcal{E}, \mathcal{N}, \mathcal{R})$ to jointly model structured and unstructured data. The core processing unit consists of three key components: (1) a graph encoder \cite{zhugmae,hou2023graphmae2} that utilizes a graph transformer to capture structural dependencies; (2) a user input module that processes natural language inquiries (e.g., procurement requests) via a text encoder; and (3) the LOM hybrid cluster, where a linear alignment projector maps graph features into the LOM's embedding space. This deep fusion enables the model to perform CoT reasoning over enterprise ontologies. 

\subsubsection{Training Method}
We begin by fine-tuning the LLM using instruction tuning \cite{DBLP:journals/kbs/DaiHWSJQ25} to enhance ontology-centric understanding and reasoning, and then integrate it into the large ontology model framework. The training objective is:
\begin{align}
\mathcal{L}=-\mathbb{E}_{\mathcal{G}, \mathbf{q}, \mathbf{a} \sim \mathcal{D}} \log P(\mathbf{a} \mid (\mathcal{G}, \mathbf{q}) ; \theta)
\end{align}
where $\mathcal{G}$ is the graph, $\mathbf{q}$ the query, $\mathbf{a}$ the answer, and $\theta$ the model parameters. Given $(\mathcal{G},\mathbf{q})$, this model is trained under $\theta$ to output correct $\mathbf{a}$, endowing the model with foundational graph understanding and reasoning.

During the training process, we align not only the knowledge graph data but also the ontology structure. Then we train the alignment projector and graph-token embeddings to align features between the text encoder and GNN. Specifically, we employ graph instruction alignment, where the LLM is trained to understand graph-structured data via graph-token-instruction pairs. We define two alignment datasets for this stage:

Intra-type alignment enhances understanding of tokens within a single meta-type by training the LLM to output the correct text sequence for a given graph-token sequence. The dataset is defined as:
\begin{align}
\mathcal{D}^{\text {intra }}=\{ [(\mathbf{e}_{k}, s_{i}), \ldots], [(\mathbf{c}_{k}, \mathbf{c}_{s_{i}}), \ldots] \}
\end{align}
We optimize this alignment using a next-token-prediction cross-entropy objective:
\begin{equation}
\mathcal{L}_{\text{intra}} = \mathbb{E}_{d \sim \mathcal{D}^{\text{intra}}} [\mathrm{CE}(d[0] \mid \text{LLM}(d[1]))]
\end{equation}

Inter-type alignment introduces multiple meta-types for complex heterogeneous relations, using tokens from different meta-types:
\begin{align}
\mathcal{D}^{\text {inter }}=\{ [(\mathbf{e}_{m}, s_{m}),(\mathbf{e}_{n}, s_{n}), \ldots], [(\mathbf{c}_{m}, \mathbf{c}_{s_{m}}),(\mathbf{c}_{n}, \mathbf{c}_{s_{n}}), \ldots] \}
\end{align}
Similarly, we train the LLM to predict the text sequence from the heterogeneous graph tokens:
\begin{equation}
\mathcal{L}_{\text{inter}} = \mathbb{E}_{d \sim \mathcal{D}^{\text{inter}}} [\mathrm{CE}(d[0] \mid \text{LLM}(d[1]))]
\end{equation}
where $\mathrm{CE}(\cdot)$ denotes the cross-entropy loss function. In both intra- and inter-type alignment tasks, $d[0]$ represents the target text sequence (ground truth), while $d[1]$ is the input sequence of graph tokens processed by the LLM. The graph token definitions are: $\mathbf{e}_{i}$ is the $i$-th graph token, $s_{i}$ its meta-type, and $\mathbf{c}_{i}$ the corresponding text description. This builds a foundation for understanding graph-structured data.

Finally, we perform multi-task fine-tuning based on the above alignment to enhance the model's capabilities across predictive and generative tasks. This phase utilizes diverse ontology-language instruction formats defined as:
\begin{align}
\mathcal{D}^{\text {multi }}=\{ \{(\mathbf{x}_{\text {pred }}, \mathbf{x}_{\text {reasoning }}) \mid \mathbf{x}_{\text {gen }}\}, \{\mathcal{G}^{\text {exp }} \mid \mathcal{G}^{\text {skg }}\}, \mathbf{t}_{i}, \mathbf{a}_{i} \}
\end{align}
The model is optimized to generate the target answer $\mathbf{a}_i$ given the instruction $\mathbf{t}_i$ and graph context $\mathcal{G}$:
\begin{equation}
\mathcal{L}_{\text{multi}} = -\mathbb{E}_{(\mathbf{t}, \mathbf{a}, \mathcal{G}) \sim \mathcal{D}^{\text{multi}}} \log P(\mathbf{a} \mid \mathbf{t}, \mathcal{G}; \theta)
\end{equation}
where $P(\mathbf{a} \mid \mathbf{t}, \mathcal{G}; \theta)$ represents the conditional probability of generating the correct answer $\mathbf{a}$ given the instruction $\mathbf{t}$ and graph context $\mathcal{G}$, parameterized by the model weights $\theta$. The input features $\mathbf{x}_{\text{pred}}$, $\mathbf{x}_{\text{reasoning}}$, and $\mathbf{x}_{\text{gen}}$ correspond to predictive, reasoning, and generative tasks, respectively. $\mathcal{G}^{\text{exp}}$ denotes explicit graph topology, while $\mathcal{G}^{\text{skg}}$ refers to the schema-enhanced knowledge graph structure. The pair $(\mathbf{t}_{i}, \mathbf{a}_{i})$ corresponds to the task-specific instruction and the ground-truth answer.

To ensure stable convergence and effective knowledge acquisition, we adopt a curriculum learning strategy that organizes training samples by difficulty. The model is first exposed to simpler predictive tasks before progressing to complex generative reasoning scenarios, thereby systematically building its capability to handle diverse graph-instruction pairs.

\subsubsection{Dataset Construction}
Existing graph-task datasets primarily enable LLMs to learn simple reasoning tasks, while complex graph reasoning tasks (e.g., minimum spanning tree, PageRank) remain difficult to learn. To address this, we introduce a CoT-enhanced dataset for learning graph reasoning. The dataset includes problem descriptions, chain-of-thought reasoning, and final answers, enabling LLMs to learn the reasoning processes of complex reasoning tasks and improve performance on graph tasks.

We further design trajectory-generation problems for reasoning tasks via a CoT-enhanced trace generator implementing four core reasoning tasks: Dijkstra shortest path, Kahn topological sorting, Prim minimum spanning tree, and predecessor node search. A stepwise recording system captures intermediate states during execution, including initialization, node visitation, and state updates, and produces natural language chain-of-thought explanations. This converts complex reasoning execution into structured training data, allowing LLMs to learn full reasoning logic rather than memorizing answers.

The final dataset contains 115k training samples spanning 19 graph reasoning tasks, from simple traversals (BFS, DFS) to complex reasoning tasks (shortest path, MST, PageRank). Data are split into two stages: 20k samples for foundational graph language model training and 95k samples for deep fusion training. We additionally build 190 evaluation samples (10 per task) for performance assessment. All data are stored in PyG format, including node feature encodings (sentence-transformer, 768-dim) and graph structure, supporting both heterogeneous and homogeneous graph training.

\subsection{Ontology Construction}
\subsubsection{Ontology Construction from Structured Data}
To address the semantic impedance mismatch and scalability challenges inherent in converting heterogeneous relational databases into ontologies, we propose a layered, decoupled, and incrementally evolving construction framework. We adopt a two-stage scanning strategy to build the ontology $\mathcal{G}_{S}=(\mathcal{V}_{S}, \mathcal{E}_{S})$. First, we extract schemas from the source tables $\mathcal{T}=\{T_1, \dots, T_n\}$ to identify explicit keys (e.g., fields containing 'id' or 'code'). To complement this, we utilize LOMs to detect implicit associations, capturing semantic variants like 'number' that elude simple keyword matching. Second, we perform a sliding-window deep scan to infer implicit foreign keys, utilizing a random sample of 1,000 records to assess value overlap. We define a multi-factor confidence function $\Phi(\cdot)$ to quantify the relationship between columns $c \in T_i$ and $c' \in T_j$:
\begin{equation}
\Phi(c, c') = \lambda_1 \mathcal{S}_{\text{name}}(c, c') + \lambda_2 \mathbb{I}_{\text{type}}(c, c') + \lambda_3 \mathcal{S}_{\text{overlap}}(c, c') + \lambda_4 \mathcal{S}_{\text{card}}(c, c')
\end{equation}
where $\mathcal{S}_{\text{name}}$ measures semantic similarity, $\mathbb{I}_{\text{type}}$ ensures data type compatibility, $\mathcal{S}_{\text{overlap}}$ quantifies value distribution overlap, and $\mathcal{S}_{\text{card}}$ infers cardinality patterns (e.g., one-to-many). An edge is established if $\Phi > \delta$. We construct a dual-layer ontology: the inter-table layer forms a graph via the above method, refined by $k$-core decomposition to retain the topological backbone, while tables are categorized top-down to abstract higher-level concepts. Finally, instance nodes are connected to their corresponding table nodes to bridge data and concepts. 

\subsubsection{Ontology Construction from Unstructured Data}
To harness the rich enterprise knowledge embedded in unstructured text, we construct the ontology $\mathcal{G}_{U}=(\mathcal{V}_{U}, \mathcal{E}_{U})$ using a systematic pipeline comprising document parsing, triple extraction \cite{zhu2023pre}, entity disambiguation, and graph construction. Specifically, during the extraction phase, we deploy a locally hosted vLLM-accelerated Qwen3 \cite{yang2025qwen3} model (temperature set to 0.1) to process documents segmented into 2,000-token chunks with a 150-token overlap. This model is tasked with extracting entities along with their descriptive attributes and identifying relationships from the set \{MENTIONS, RELATES\_TO, IS\_A\}.

To resolve non-standard entity mentions and ambiguity, we apply layered merging rules: (1) initial surface matching via edit distance ($\ge 0.85$) and substring containment ($\ge 60\%$); (2) domain-specific normalization; and (3) deep semantic matching using BGE-M3 \cite{chen2024bge} embedding cosine similarity ($\ge 0.85$). This ensures synonymous mentions are consolidated into canonical nodes in $\mathcal{V}_{U}$.

The unified heterogeneous graph $\mathcal{G} = \mathcal{G}_{S} \cup \mathcal{G}_{U}$ is synthesized by integrating these two layers. We employ a heuristic tagging algorithm to generate attribute tags for file nodes in $\mathcal{G}_{U}$, which are subsequently matched against descriptions in the instance layer of $\mathcal{G}_{S}$ to establish semantic associations $\Psi: \mathcal{V}_{U} \to \mathcal{V}_{S}$, effectively bridging the semantic gap between raw text and relational data. Furthermore, we utilize the link prediction capabilities of the LOM to infer missing relationships, thereby densifying and completing the graph structure.

\section{Experiments}
\subsection{Dataset}
We develop a graph reasoning dataset for training and evaluation, which encompasses 19 diverse tasks across six categories: traversal, graph properties, node similarity, paths and flows, centrality, and tree structures. The data, stored in PyG format, incorporates 768-dimensional node features encoded by sentence-transformers and detailed edge structures. The training corpus consists of 95,000 samples supporting multi-stage instruction tuning, while the evaluation benchmark comprises 190 samples (10 per task type) stratified by difficulty: simple (e.g., BFS, node degree), medium (e.g., connectivity), and difficult (e.g., shortest path, PageRank). The graphs vary in size from 5 to 50 nodes and 5 to 200 edges, covering undirected, directed, and weighted types.

\subsection{Hyper-parameters}
We adopt Qwen3-4B-Instruct as the backbone LLM, featuring 4B parameters, a hidden size of 2,560, 36 layers, 32 attention heads, and a vocabulary of 151,936, with the maximum generation length set to 4,096 tokens. To capture graph structural information, we utilize a pre-trained graph transformer encoder consisting of 3 layers with a hidden dimension of 128 and 8 attention heads. Node semantics are initialized using the all-mpnet-base-v2 Sentence-Transformer to generate 768-dimensional features. These features are processed by the graph encoder via dimension-matching projection layers and finally mapped to the LLM’s 2,560-dimensional embedding space through a linear graph projector.

\subsection{Evaluation Metrics}
We evaluate performance using accuracy as the primary metric on a benchmark comprising 190 test samples, evenly distributed across 19 distinct graph reasoning tasks. These tasks span six categories: traversal (BFS, DFS), graph properties (node degree, neighbor query, edge existence, connectivity, cycle detection, bipartiteness, connected components, diameter), node similarity (common neighbors, Jaccard similarity), paths and flows (predecessor, topological sort, shortest path, maximum flow), centrality (PageRank, clustering coefficient), and tree structure (minimum spanning tree). For assessment, we apply strict task-specific correctness criteria: exact order for traversals, precise values for numerical tasks, correct classifications for boolean queries, set matching for list outputs, and validity checks for path-related tasks.

\subsection{Main Results}


As shown in Figure \ref{fig.ranking}, LOM-4B demonstrates superior performance across the benchmark, achieving the top rank with an overall accuracy of 89.47\% (170/190 correct). This represents a substantial improvement over competing approaches, validating the effectiveness of our proposed method in handling complex graph reasoning tasks. The model's architecture, which integrates a graph encoder with a powerful large language model, evidently provides a distinct advantage in interpreting and solving graph-structured problems.

The comparison further highlights the performance gap between specialized and general-purpose models. The second-tier models, including DeepSeek-V3.2 \cite{liu2025deepseek}, GraphInstruct \cite{DBLP:journals/corr/abs-2403-04483}, and Doubao-1.8, achieve comparable results in the range of 78.42\% to 79.47\%, yet they trail LOM-4B by approximately 10\%. Notably, the base Qwen3 models (Qwen-Max and Qwen3-4B-Instruct) exhibit significantly lower accuracies of 57.37\% and 18.95\% respectively, underscoring the necessity of targeted instruction tuning and structural encoding for effective graph reasoning execution.



\begin{center}
\includegraphics[width=0.45\textwidth]{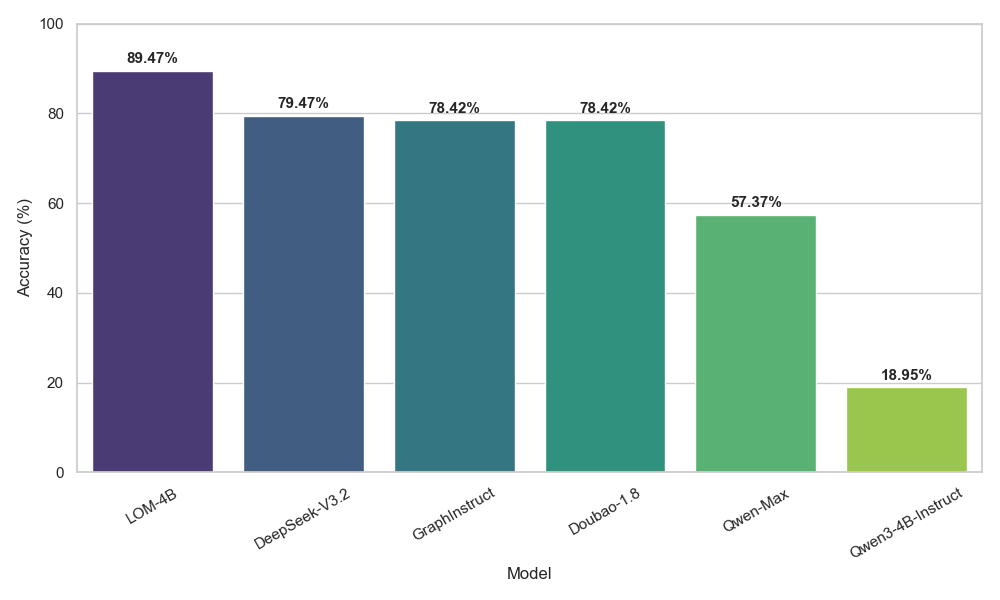}
\captionof{figure}{Model Performance Ranking}
\label{fig.ranking}
\end{center}

\subsection{Analysis}

\begin{figure*}[!htbp]
\centering
\includegraphics[width=\textwidth]{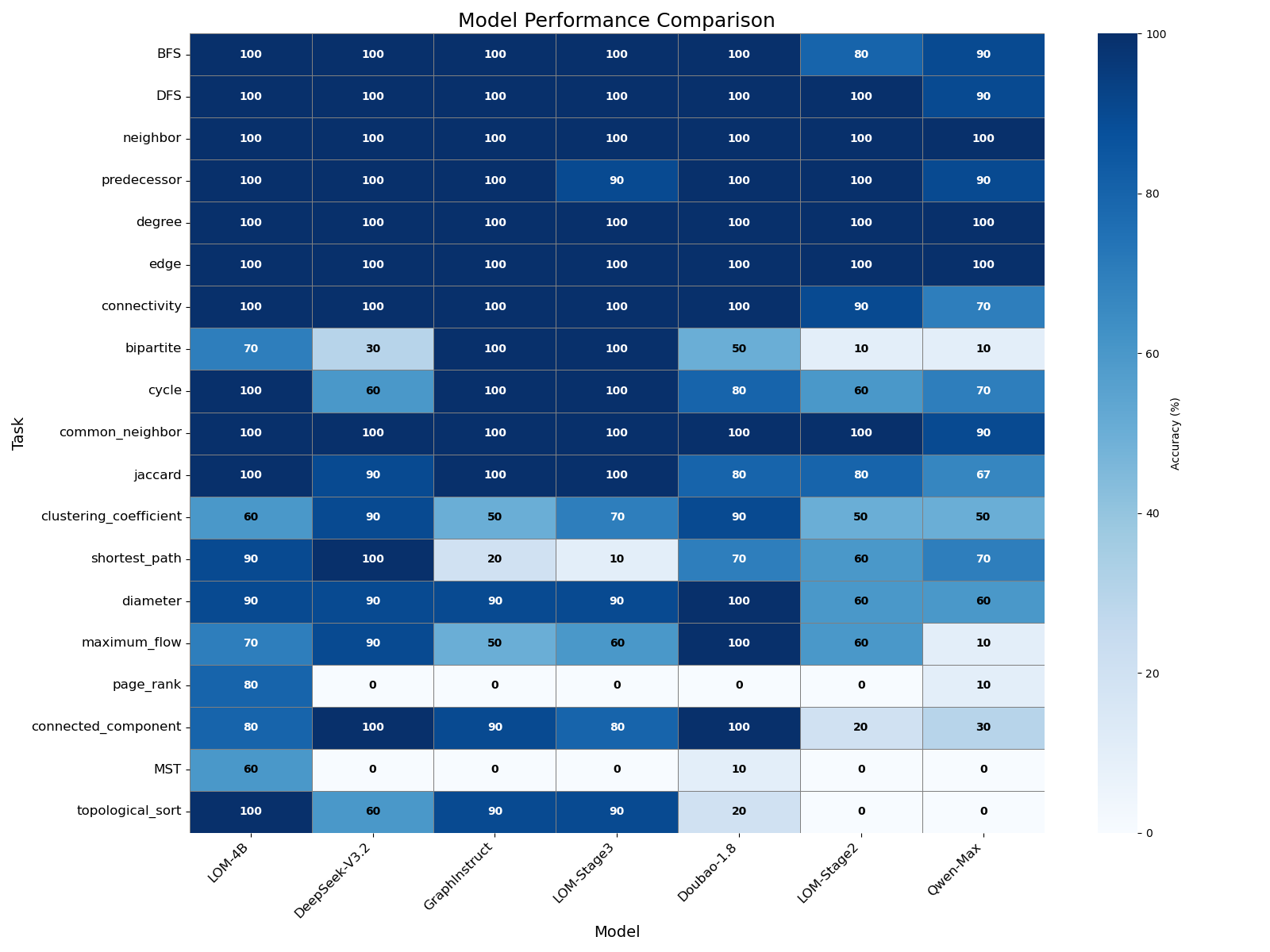}
\caption{Model Performance Heatmap}
\label{fig.heatmap}
\end{figure*}

As shown in Figure \ref{fig.heatmap}, we present a detailed breakdown of model performance across 19 specific graph tasks. On fundamental graph traversal and property retrieval tasks---such as BFS, DFS, neighbor queries, and degree calculation---most advanced models, including LOM-4B, DeepSeek-V3.2, and GraphInstruct, achieve near-perfect accuracy (100\%). This indicates that current LLMs, when equipped with appropriate training or structural awareness, have successfully mastered the basic syntax and local connectivity rules of graph data.

However, a significant divergence in performance emerges on computationally intensive and global reasoning tasks. LOM-4B exhibits remarkable robustness on algorithmic challenges that baffle other models. Specifically, for PageRank and minimum spanning tree (MST) tasks, where other leading models like DeepSeek-V3.2 and GraphInstruct score near 0\%, LOM-4B maintains high accuracy (80\% and 60\% respectively). Similarly, in topological sorting, LOM-4B achieves 100\% accuracy, outperforming the next best competitor by a margin, demonstrating its superior capacity to reason about global graph structures and dependencies.

The results also shed light on the impact of training stages and model specialization. Comparing LOM-Stage2, LOM-Stage3, and the final LOM-4B reveals a clear progression in capability, particularly for complex tasks like bipartite matching and shortest paths. While general-purpose models like Doubao-1.8 and DeepSeek-V3.2 show sporadic strengths (e.g., 100\% in maximum flow or shortest path), they lack the consistent versatility of LOM-4B. The contrast with Qwen-Max further emphasizes that scale alone is insufficient for graph reasoning; the specialized architecture and instruction tuning of LOM are critical for bridging the gap between simple retrieval and complex algorithmic execution.


\subsection{Ablation Study}

As shown in Figure \ref{tab:ablation}, we conduct an ablation study to quantify the contribution of each key component in LOM-4B. The full model achieves the highest accuracy of 89.47\%. Removing the CoT reasoning mechanism leads to a significant drop of over 11\%, reducing accuracy to 78.42\%. This indicates that explicit reasoning steps are crucial for guiding the model through complex multi-step graph reasoning tasks, preventing it from jumping to incorrect conclusions and ensuring more robust problem-solving strategies.

The impact of instruction tuning and structural encoding is even more profound. Without instruction tuning (``w/o Instruct''), performance further degrades to 61.66\%, highlighting the importance of aligning the LLM with specific graph-related tasks. Most strikingly, the baseline model without both the graph encoder and instruction tuning (``w/o GNN, Instruct'') collapses to a mere 18.95\%. This drastic decline confirms that standard LLMs, lacking specific structural embeddings and task adaptation, are fundamentally ill-equipped for graph reasoning, validating the necessity of our proposed hybrid architecture.

\begin{table}[ht]
\centering
\caption{Ablation Study Results}
\label{tab:ablation}
\begin{tabular}{l c}
\hline
Settings & Accuracy \\
\hline
LOM-4B & 89.47\% \\
w/o CoT & 78.42\% \\
w/o Instruct & 61.66\% \\
w/o GNN, Instruct & 18.95\% \\
\hline
\end{tabular}
\end{table}

\section{Conclusion}
We have presented the LOM, a unified framework for enterprise ontology construction and reasoning that effectively bridges the gap between structured databases and unstructured textual knowledge. By integrating a dual-layer ontology construction method with a three-stage instruction alignment pipeline—spanning ontology instruction fine-tuning, text-ontology grounding, and multi-task instruction tuning—we enable the model to perform complex, structure-aware reasoning over heterogeneous enterprise data. Our experiments demonstrate that the LOM-4B achieves state-of-the-art performance on diverse ontology reasoning tasks, proving its efficacy in handling the intricacies of real-world enterprise environments.

In future work, we plan to enhance the performance of LOM on complex ontology reasoning tasks, particularly focusing on challenging tasks like MST and PageRank where current models consistently achieve low scores. We plan to adopt more reinforcement learning strategies to enable deeper structural understanding and more accurate multi-hop inference to overcome these limitations.

\printcredits

\bibliographystyle{cas-model2-names}

\bibliography{cas-refs}

\end{document}